\documentclass[]{bytedance_seed}

\usepackage[toc,page,header]{appendix}

\usepackage{minitoc}
\usepackage{booktabs}
\usepackage{makecell}
\usepackage{fix-cm}

\usepackage{amsmath,amsfonts,bm}

\def\eqref#1{equation~\ref{#1}}

\def\1{\bm{1}}

\DeclareMathAlphabet{\mathsfit}{\encodingdefault}{\sfdefault}{m}{sl}
\SetMathAlphabet{\mathsfit}{bold}{\encodingdefault}{\sfdefault}{bx}{n}

\usepackage{colortbl}
\usepackage{xcolor}
\definecolor{mygray}{gray}{.9}
\definecolor{goldenrod}{RGB}{245,245,220}
\newlength\savewidth
\newcolumntype{a}{>{\columncolor{mygray}}c}
\usepackage{fontawesome5}
\usepackage{booktabs}
\usepackage{makecell}
\usepackage{fontawesome5}
\usepackage{lipsum}
\usepackage{comment}
\usepackage{multirow}
\usepackage{wrapfig}
\usepackage{algorithm}
\usepackage{algorithmic}
\usepackage{xfrac}  

\usepackage{graphicx}
\usepackage{color}

\usepackage{amsthm}
\definecolor{darkgreen}{rgb}{0,0.7,0}

\definecolor{mygraytext}{gray}{.75}

\title{IOI: Decoupling Kinematics and Physics for Interactive World Models}

\author[1,2,\star]{Chengyu Bai}
\author[1,2,\star,\bullet]{Peidong Jia}
\author[1,2,\star]{Tiecheng Guo}
\author[1,2]{Yu-Kai Wang}
\author[2]{Rui Ma}
\author[2]{Fangyuan Zhao}
\author[1,2]{Chun-Kai Fan}
\author[2]{Xiaobao Wei}
\author[1,2]{Jintao Chen}
\author[2]{Hao Wang}
\author[2]{Ying Li}
\author[1]{Xiaozhu Ju}
\author[1,\dagger]{Jian Tang}
\author[2,\dagger]{Shanghang Zhang}

\affiliation[1]{Beijing Innovation Center of Humanoid Robotics}
\affiliation[2]{Peking University}

\contribution[\star]{These authors contributed equally}
\contribution[\bullet]{Project leaders}
\contribution[\dagger]{Corresponding author}

\definecolor{highlight}{RGB}{220,230,241}
\definecolor{badcase}{RGB}{255,200,200}

\abstract{
Developing generalist embodied agents demands scalable and interactive environments that provide both visually realistic feedback and accurate action-conditioned dynamics for policy learning and evaluation. Interactive world models offer a promising paradigm to address this challenge by simulating complex action-conditioned dynamics. However, existing methods often struggle to simultaneously ensure precise control alignment and physically plausible visual feedback, primarily stemming from the absence of explicit structural constraints in purely data-driven frameworks. To address this limitation, we propose IOI, a hybrid interactive world model that integrates analytical kinematic priors with learned physical dynamics. Unlike purely data-driven generation approaches prone to spatiotemporal drift, IOI introduces explicit kinematic guidance, computing forward kinematics from action sequences to obtain accurate motion trajectories. These trajectories are rendered into synchronized front, side, and top orthographic projections, which eliminate the need for extrinsic camera calibration to align rendered motions with observed scenes. A lightweight yet effective Multi-view Kinematic Aggregation and Injection module then fuses these multi-view geometric cues and injects them into the video generator, providing geometry-consistent and view-invariant guidance. By conditioning video generation on deterministic kinematic trajectories, IOI establishes a principled synergy between the analytical simulator and the world model. By decoupling deterministic embodiment motion into the kinematic prior, the generator's representational capacity is fully dedicated to modeling stochastic physical interactions and environmental dynamics. 
Extensive experiments on the RoboTwin benchmark comprehensively validate IOI across three critical dimensions: kinematic fidelity, out-of-distribution (OOD) generalization, and policy evaluation. IOI achieves state-of-the-art simulation performance, while demonstrating robust zero-shot generalization to unseen OOD tasks. Furthermore, IOI serves as a highly reliable policy evaluator, yield success rates that closely align with ground-truth physics simulators. On real-world platforms, policies trained on IOI-synthesized data match those trained on teleoperation demonstrations, solidifying its practical value for embodied policy learning.
}

\date{May 10, 2026}
\correspondence{Shanghang Zhang at \email{shanghang@pku.edu.cn}}

\begin{document}
\maketitle

\begin{figure}
    \centering  
    \captionsetup{type=figure}
    \includegraphics[width=\textwidth]{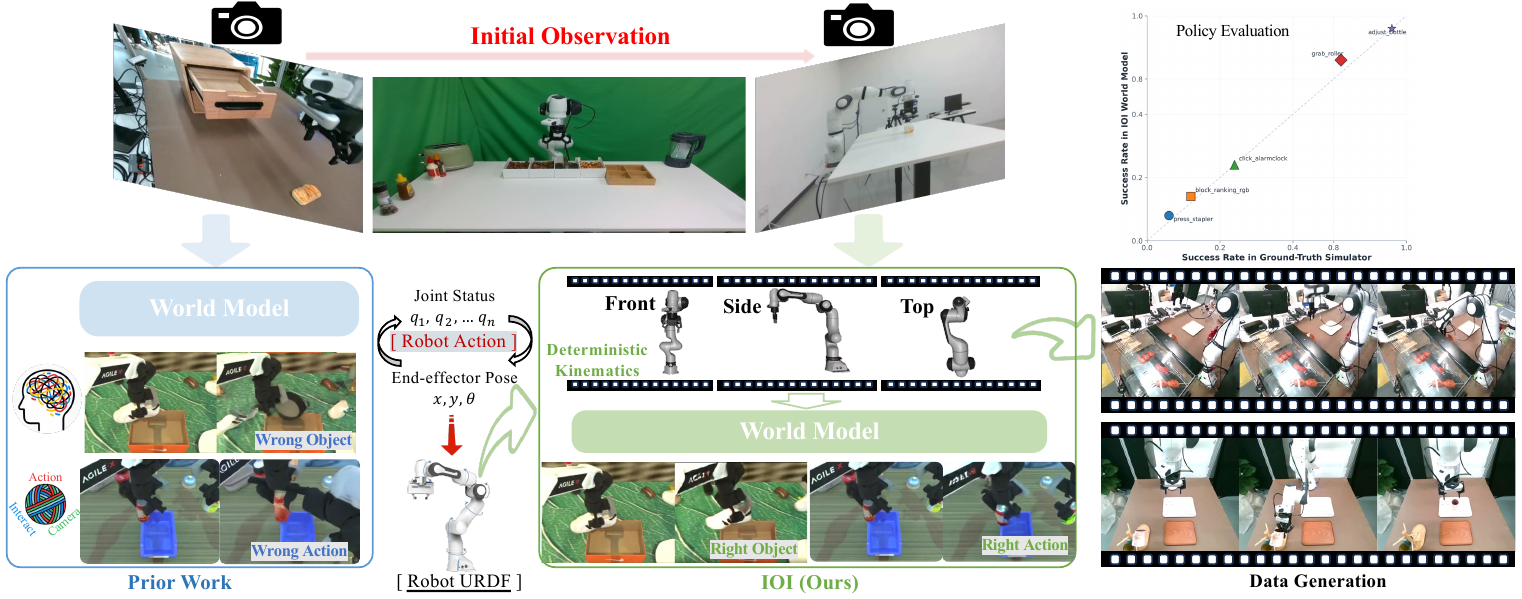}
    \captionof{figure}{The illustration of UniEdit-I. We introduce a novel training-free framework named UniEdit-I to enable the unified VLM with image editing capability via three iterative steps: \textbf{understanding, editing, and verifying}.}
    \label{fig:teaser}  
\end{figure}

\section{Introduction}
\label{sec:intro}
Developing generalist embodied agents requires scalable, visually realistic, and interactive environments for policy learning and evaluation~\cite{brohan2022rt1,brohan2023rt2,black2024pi0,kim2024openvla, li2026manipdreamer3d, chi2025wow}.
Traditionally, the field relied on physics-based simulators, which require labor-intensive asset creation and operate under simplified physical rules, resulting in a significant sim-to-real gap in both visual appearance and contact dynamics~\cite{aljalbout2025reality, nasiriany2024robocasa, tao2024maniskill3}.
Real-world teleoperation (e.g., ALOHA~2~\cite{aldaco2024aloha2}) offers an alternative by capturing authentic physical interactions, yet only provides offline trajectories rather than an sandbox for agent exploration~\cite{gu2025emma}.

To bridge the gap between unrealistic simulators and static offline teleoperation datasets, interactive world models have emerged as a promising paradigm~\cite{yang2024unisim, zhu2025irasim, jang2025dreamgen, guo2026ctrlworld}.
By framing environmental dynamics as an action-conditioned video generation task, they learn visual transitions directly from real-world observations, bypassing the need for explicitly programmed physical rules.
This enables agents to explore counterfactual action sequences in a photorealistic and interactive environment, eliminating the overhead of physical asset construction and human demonstration collection.

Despite this transformative potential, current interactive world models operate predominantly under a purely data-driven paradigm. By exclusively mapping visual observations and actions to future video frames~\cite{yang2024unisim, zhu2025irasim, huang2025vid2world}, they rely entirely on pixel-level statistical correlations. This fundamental lack of explicit physical grounding makes them susceptible to two critical failure modes, \textit{action deviation} and \textit{state implausibility}, as illustrated in Figure~\ref{fig:teaser}. At the action level, accumulated prediction errors cause the synthesized motion to progressively diverge from the intended control instructions. At the state level, models fail to maintain physically consistent object states, producing unrealistic deformations, geometric penetrations, and violations of object permanence~\cite{zhen2025tesseract, blattmann2023svd}. Consequently, these persistent visual artifacts and dynamic violations render the generated trajectories fundamentally unreliable for robust policy learning and evaluation.

To resolve the fundamental dilemma between enforcing precise geometric control and generating stochastic visual feedback, we propose \textbf{IOI}, a novel interactive world model that explicitly decouples deterministic robot kinematics from complex environmental dynamics. By integrating analytic priors, our approach establishes a collaborative framework where a physical simulator governs exact structural motion, leaving the neural generator to exclusively render interactive visual transitions.

Specifically, we utilize the Universal Robot Description Format (URDF) to analytically map the input joint states into precise 3D articulated structures. To circumvent the severe noise and inaccuracies introduced by traditional extrinsic camera calculations, we apply an orthogonal projection mechanism to translate these 3D models into the 2D visual domain. The projected geometric features are subsequently processed by the Multiview Kinematic Aggregation and Injection (MKAI) module, which consists of three dedicated components. A multi-view fusion module first aggregates the front, side, and top orthographic projections via lightweight MLPs into a unified geometric representation. A cross-attention mechanism then aligns this representation with the visual observation, grounding the kinematic prior in the observed scene context. Finally, a multi-layer VACE-based injection module incorporates the URDF-conditioned structural prior into the video generator, enabling geometry-consistent and physically grounded video synthesis. By introducing dedicated conditioning branches, we explicitly guide the generative process using deterministic structural constraints. This design tightly aligns precise robot control signals with physically plausible visual interaction synthesis.

We comprehensively evaluate IOI across the RoboTwin 2.0 simulator and physical robotic platforms, rigorously assessing its kinematic fidelity, out-of-distribution (OOD) generalization, and utility as a policy evaluator. IOI establishes a new state of the art against existing methods. Specifically, for standard manipulation tasks, IOI yields significant improvements over prior methods across all standard evaluation metrics. Furthermore, it demonstrates robust zero-shot generalization to unseen OOD tasks, significantly outperforming baselines in temporal coherence and perceptual quality. Beyond generation quality, IOI serves as a highly reliable surrogate for embodied policy evaluation, yielding success rates that closely align with ground-truth physics simulators. Finally, in real-world deployments, policies trained exclusively on IOI-synthesized data achieve success rates comparable to those trained on teleoperation demonstrations, validating IOI as an effective data generation engine for embodied policy learning.

Our contributions are as follows:

\begin{itemize}
    \item \textbf{A Novel Paradigm Shift.} We propose IOI, an interactive world model that integrates analytical kinematic priors with learned physical dynamics, explicitly decoupling deterministic embodiment motion from stochastic environmental interactions. This principled design substantially mitigates action deviation and state implausibility, two critical failure modes inherent in purely data-driven methods.
    \item \textbf{Elegant and Effective Architecture.} We introduce an end-to-end pipeline consisting of URDF-based kinematic solver, a multi-view orthographic renderer, and a Multiview Kinematic Aggregation and Injection module, enabling geometry-consistent robotic simulation without extrinsic camera calibration.
    \item \textbf{State of the Art Physical Interaction Performance.} Extensive evaluations on both simulation benchmarks and real-world platforms demonstrate that IOI achieves superior generation fidelity, reliable policy evaluation closely matching ground-truth testing, and policies trained on IOI-synthesized data performing comparably to those trained on teleoperation demonstrations.
\end{itemize}

\section{Related Work}
\label{sec:relatedwork}

\subsection{Interactive World Models}
World models predict future states from observations and actions, serving as internal simulators for embodied agents. Early methods forecast latent representations without pixel-level reconstruction~\cite{assran2023self, bardes2023v, liu2024lwm}, while recent approaches leverage video diffusion models~\cite{wan2025wan, yang2024cogvideox, blattmann2023svd, hacohen2024ltx} to synthesize photorealistic visual rollouts. Foundation models such as Genie~\cite{bruce2024genie} and Cosmos~\cite{agarwal2025cosmos} further demonstrate interactive environment generation from visual inputs. In robotic manipulation, fine-grained action-conditioned generation is essential for interactive simulation. IRASim~\cite{zhu2025irasim} maps continuous action trajectories to video synthesis, and Ctrl-World~\cite{guo2026ctrlworld} builds a controllable world model supporting diverse action modalities. Concurrent efforts adapt frozen video models through action adapters~\cite{rigter2024avid}, synthesize multi-modal interactive rollouts~\cite{yang2024unisim}, or employ world models for direct motion planning~\cite{cen2025worldvla, bi2025motus}. Despite this progress, these models treat the robot and environment as an entangled whole, fusing continuous action embeddings with visual tokens end-to-end. This entanglement forces the network to jointly model deterministic robot geometry and stochastic environmental dynamics, which can lead to geometric inconsistencies such as self-penetration and physically implausible motions~\cite{zhen2025tesseract}.

\subsection{Physical Priors in Generative Models}
Providing generative models with explicit physical structure of the robot offers a principled way to separate kinematic modeling from environmental dynamics. ControlNet~\cite{zhang2023adding} demonstrates that injecting auxiliary spatial conditions such as depth maps and pose skeletons into diffusion models effectively guides structural generation while preserving generative capacity. Subsequent works extend this paradigm to video generation with temporal-aware conditioning~\cite{zhao2025motionprompting}, providing an architectural foundation for such condition injection. In robotics, EC-Flow~\cite{ecflow2025} predicts executable actions from action-free video via embodiment-centric optical flow, targeting policy learning rather than visual synthesis. ManipDreamer3D~\cite{li2026manipdreamer3d} projects planned 3D trajectory onto 2D panel as generative priors. BridgeV2W~\cite{bridgev2w2025} injects URDF-rendered embodiment masks into a pretrained video model via ControlNet-style conditioning, yet its reliance on extrinsic camera parameters for mask rendering introduces viewpoint sensitivity and limits deployability across diverse camera configurations. In contrast, we exploit the asymmetry between deterministic robot kinematics and stochastic scene dynamics more thoroughly by projecting URDF-derived articulated structures via viewpoint-invariant orthographic projections and injecting them through a dedicated module, enabling the diffusion model to focus its capacity entirely on environmental dynamics.

\section{Problem Formulation}

\begin{figure*}[t]
  \centering
  \includegraphics[width=\textwidth]{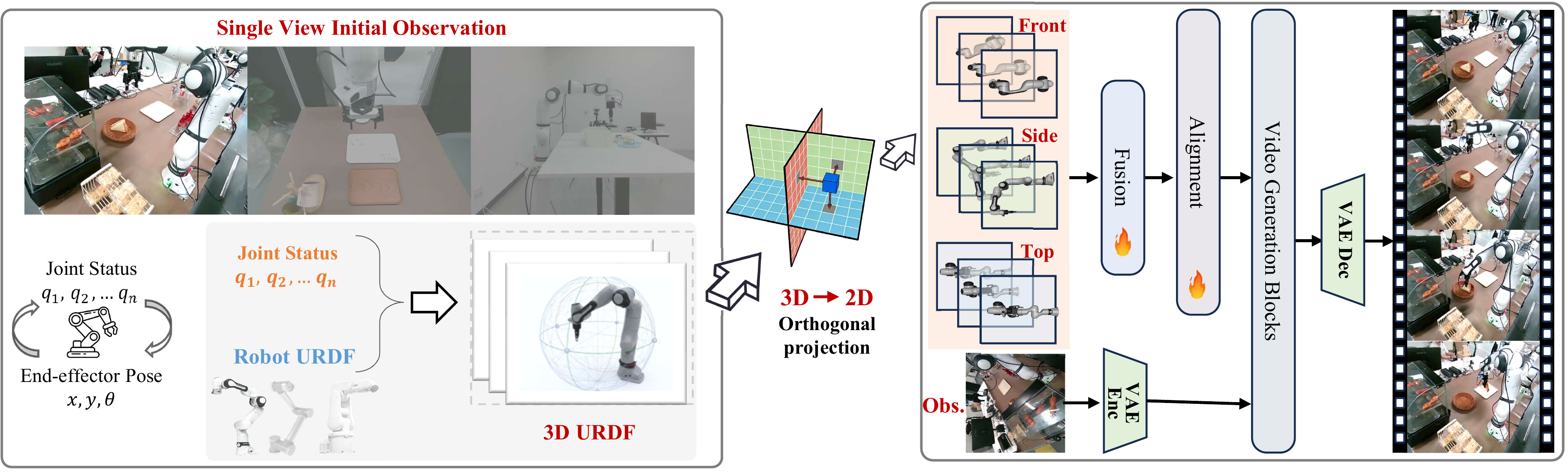}
  \caption{Pipeline Overview. IOI adopts a hybrid paradigm that analytically computes deterministic robot kinematics from the URDF model and action sequences, generating multi-view 2D geometric conditions. These kinematic priors are aggregated and injected into a diffusion-based video generator via the proposed MKAI module, providing accurate geometric guidance for learning stochastic scene dynamics without implicitly inferring robot embodiment from low-dimensional actions.}
  \label{fig:pipeline}
\end{figure*}

This work focuses on constructing an interactive world model that serves as a photorealistic and explorable imagined environment for embodied manipulation policy learning and evaluation.
Following the prevailing paradigm of action-conditioned video generation~\cite{zhu2025irasim, jang2025dreamgen}, our formulation is grounded in the observation that robot ego-motion is deterministic under the kinematic chain, whereas contact outcomes and scene evolution are inherently stochastic due to the uncertainty in robot-environment interactions.
This distinction motivates a factorized formulation that separates deterministic embodiment motion from stochastic interaction dynamics.

Specifically, given $V_{t-h:t}=\{I_{t-h},\dots,I_t\}$ as a history of $h$ observation frames, $T$ future actions $A_{t+1:t+T}=\{a_{t+1},\dots,a_{t+T}\}$, and the robot embodiment defined by its URDF model $\mathcal{M}$, our objective is to synthesize a future video clip $\hat{V}_{t+1:t+T}=\{\hat{I}_{t+1},\dots,\hat{I}_{t+T}\}$. This prediction must accurately capture both the robot motion driven by the action sequence and the resulting scene dynamics arising from robot--environment interactions. Formally, we aim to model the conditional distribution:
\begin{equation}
    p(\hat{V}_{t+1:t+T}\mid V_{t-h:t}, A_{t+1:t+T}, \mathcal{M})
\end{equation}
Here, $\mathcal{M}$ introduces no additional assumption, as URDF models are commonly available in manipulation pipelines that involve motion planning or Cartesian-space control~\cite{brohan2023rt2,black2024pi0}.

Since $\mathcal{M}$ and $A_{t+1:t+T}$ together fully determine the robot trajectory via forward kinematics, deterministic embodiment motion can be resolved analytically without burdening the generator.
This frees the generator to focus its representational capacity on stochastic physical interactions and scene changes. 

\section{Method}

Our framework, \textbf{IOI}, employs a hybrid modeling paradigm that analytically resolves deterministic robot kinematics while learning stochastic scene dynamics through conditional video generation. Given the action sequence $A_{t+1:t+T}$ and the URDF model $\mathcal{M}$, IOI first computes forward kinematics to obtain 3D articulated motion trajectories, which are then projected into synchronized front, side, and top orthographic views as lightweight 2D geometric conditions described in Sec.~\ref{sec:kinematic}. These multi-view kinematic representations are subsequently encoded, fused, and injected into the diffusion backbone by a dedicated Multi-view Kinematic Aggregation and Injection module, detailed in Sec.~\ref{sec:mkai}. By conditioning the generator on explicit kinematic structure rather than raw low-dimensional actions, IOI eliminates the need to implicitly learn embodiment geometry from data, providing precise and generalizable guidance for geometry-consistent video synthesis.

\subsection{Kinematic Prior Extraction}
\label{sec:kinematic}
IOI begins by analytically deriving an explicit kinematic trajectory from the input action sequence, grounded in the robot's URDF model $\mathcal{M}$. Formally, the URDF is represented as $\mathcal{M}=\{\mathcal{L}, \mathcal{J}, \Theta\}$, where $\mathcal{L}$ denotes links set, $\mathcal{J}$ joints set, and $\Theta$ the kinematic parameters including joint axes, limits, and parent-child transforms.

Given an action sequence $A_{t+1:t+T} = \{a_{t+1}, \dots, a_{t+T}\}$, we first convert it into a unified joint-space trajectory state $\{q_{t+1},\dots,q_{t+T}\}$, where the conversion depends on the action representation.

For joint-space actions where $a_k = \Delta q_k \in \mathbb{R}^n$ denotes the joint displacement and $n = |\mathcal{J}|$ is the number of joints, each state is obtained via incremental integration:
\begin{equation}
    q_k = q_{k-1} + \Delta q_k,
\end{equation}
For Cartesian end-effector actions where $a_k \in \mathrm{SE}(3)$ denotes the target end-effector pose, we instead solve inverse kinematics using the previous joint state as initialization to ensure temporal coherence:
\begin{equation}
    q_k = \mathrm{IK}(a_k, q_{k-1};\, \mathcal{M}).
\end{equation}

Given the joint-space trajectory $\{q_{t+1},\dots,q_{t+T}\}$, we apply forward kinematics to obtain the 3D pose of each link at every timestep:
\begin{equation}
    \{T_{l,k}\}_{l=1}^{|\mathcal{L}|} = \mathrm{FK}(q_k;\, \mathcal{M}),
\end{equation}
where $T_{l,k} \in \mathrm{SE}(3)$ denotes the rigid transformation of link $l$ at timestep $k$, and $|\mathcal{L}|$ is the total number of links. The complete robot geometry at timestep $k$ is reconstructed by applying each link transformation to its canonical mesh $P_l \subset \mathbb{R}^3$:
\begin{equation}
    G_k = \bigcup_{l=1}^{|\mathcal{L}|} T_{l,k}\, P_l,
\end{equation}
where $G_k$ represents the full articulated robot geometry in world coordinates. $G_k$ serves as the input to the subsequent orthographic projection stage, which renders view-consistent 2D geometric conditions without requiring extrinsic camera calibration. This stage analytically lifts the low-dimensional action sequence into an explicit spatiotemporal robot trajectory, offloading deterministic kinematic reasoning from the generative model and freeing it to focus exclusively on modeling stochastic scene dynamics.

Directly projecting the reconstructed 3D robot geometry $G_k$ into the observation camera frame and injecting it into the video generator requires precise extrinsic camera parameters to ensure alignment between the rendered geometry and the observed scene. However, accurate extrinsic estimation is inherently impractical in real-world deployment due to occlusions, dynamic lighting, and calibration drift, making perspective-based rendering error-prone.

To overcome these limitations, IOI re-parameterizes the kinematic trajectory $G_k$ into synchronized orthographic renderings from three canonical views, namely front, side, and top. This canonical representation decouples the control signal from the observation camera pose, eliminating the need for explicit extrinsic calibration between the rendering coordinate system and the input camera.

Formally, we define three orthographic projection operators \(\Pi^{f}\), \(\Pi^{s}\), and \(\Pi^{t}\) corresponding to the front, side, and top views, respectively. For each time step \(k\), the articulated robot geometry \(G_k\) is rendered under these three projections:
\begin{equation}
R_k^{v}=\mathcal{R}(G_k;\Pi^{v}), \quad v\in\{f,s,t\},
\end{equation}
where $\mathcal{R}(\cdot)$ denotes the rasterization-based renderer and $R_k^v \in \mathbb{R}^{H \times W \times C}$ is the resulting 2D rendered image of the robot geometry at timestep $k$ under view $v$, with $H$, $W$, and $C$ denoting height, width, and channel dimensions respectively. Stacking all time steps yields three synchronized kinematic control sequences:
\begin{equation}
R^{v}_{t+1:t+T}=\{R_{t+1}^{v},\dots,R_{t+T}^{v}\}, \quad v\in\{f,s,t\}.
\end{equation}

These orthographic renderings bridge analytical kinematics with learned video dynamics by providing an explicit yet compact geometric intermediate representation. Unlike perspective projection, which introduces depth-dependent scale distortion and requires precise camera calibration, orthographic projection preserves true geometric proportions regardless of depth. More importantly, the three canonical views are geometrically complementary, where the front view captures lateral motion, the side view captures depth-wise motion, and the top view captures planar motion, together providing complete coverage of 3D robot motion without redundancy. This lossless decomposition of 3D geometry into three 2D projections yields a view-invariant representation that remains consistent across arbitrary observation cameras.

\subsection{Multi-view Kinematic Aggregation and Injection}
\label{sec:mkai}

\begin{figure}[th]
  \centering
  \includegraphics[width=0.5\linewidth]{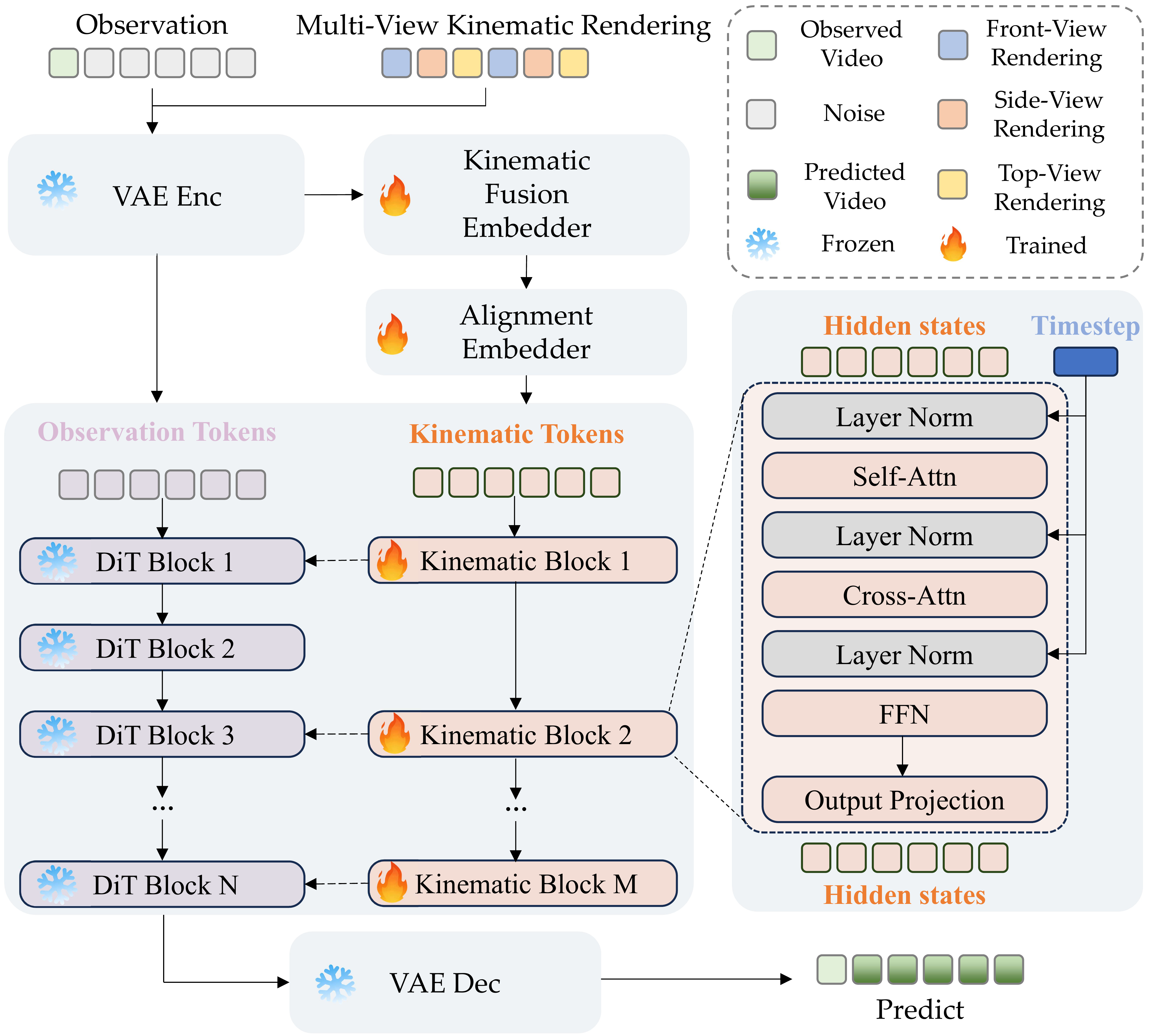}
  \caption{MKAI Overview. MKAI encodes multi-view kinematic renderings, aggregates cross-view spatial-temporal features, and hierarchically injects structured kinematic conditions into the diffusion backbone via lightweight adapters for accurate and consistent video generation.}
  \label{fig:model}
\end{figure}

Tri-view orthographic renderings offer complementary geometric cues, yet leveraging them effectively requires a dedicated architecture. Specifically, such an architecture must encode view-specific signals, aggregate them into a unified geometric representation, and inject the resulting kinematic prior into the diffusion backbone without compromising its pretrained generative capacity. To this end, we propose the \textbf{Multi-view Kinematic Aggregation and Injection (MKAI)} module as shown in Figure~\ref{fig:model}, which consists of three components, namely a Kinematic Fusion Embedder, an Alignment Embedder, and a series of Kinematic Blocks that hierarchically inject geometric guidance into the frozen DiT backbone.

\subsubsection{Kinematic Fusion Embedder}
Given the frozen VAE latents $z^{v} = \mathrm{Enc}(R^{v}_{t+1:t+T})$ for 
$v \in \{f, s, t\}$, learnable temporal embeddings $E_{\mathrm{time}} \in 
\mathbb{R}^{T \times d_{\mathrm{vae}}}$ are first incorporated to preserve temporal 
ordering:
\begin{equation}
    \tilde{z}^{v} = z^{v} + E_{\mathrm{time}}, \quad v \in \{f, s, t\}.
\end{equation}

The temporally-augmented latents $\{\tilde{z}^{f}, \tilde{z}^{s}, \tilde{z}^{t}\}$ 
are then fused into a unified kinematic representation via a shared fusion encoder 
$E_{\mathrm{fusion}}$:
\begin{equation}
    C_{k} = E_{\mathrm{fusion}}(\tilde{z}^{f}, \tilde{z}^{s}, \tilde{z}^{t}),
\end{equation}
where $C_{k} \in \mathbb{R}^{T \times d}$ denotes the aggregated kinematic feature 
sequence. $E_{\mathrm{fusion}}$ is implemented as a two-layer MLP that jointly 
processes the concatenated tri-view latents, ensuring cross-view feature homogeneity 
while mitigating view-specific distribution shifts.

\subsubsection{Alignment Embedder}
The aggregated kinematic representation $C_{k}$ is subsequently tokenized into 
the hidden space of the diffusion backbone via the Alignment Embedder 
$E_{\mathrm{align}}$, which follows the same 3D convolutional patchification 
design as the VACE Context Embedder~\cite{jiang2025vace}. Concretely, $C_{k}$ 
is partitioned into spatiotemporal patches and linearly projected with 3D 
sinusoidal positional encodings:
\begin{equation}
    \mathcal{K} = E_{\mathrm{align}}(C_{k}),
\end{equation}
where $\mathcal{K} \in \mathbb{R}^{L \times d_{\mathrm{dit}}}$ denotes the 
resulting Kinematic Tokens, $L$ is the number of spatiotemporal patches, and 
$d_{\mathrm{dit}}$ is the hidden dimension of the DiT backbone. This alignment 
step ensures that the kinematic tokens share the same spatial resolution, 
temporal structure, and feature dimensionality as the video tokens in the 
main DiT branch, enabling direct additive injection in the subsequent 
Kinematic Blocks.

\subsubsection{Kinematic Blocks and Injection}
The Kinematic Tokens $\mathcal{K}$ are processed by $M$ Kinematic Blocks. 
Each Kinematic Block comprises layer normalization, self-attention over kinematic 
tokens, cross-attention with observation tokens, a feed-forward network, and an 
output projection. Formally, for the $i$-th Kinematic Block:
\begin{equation}
    \mathcal{K}_{i+1} = \mathrm{KBlock}_i(\mathcal{K}_i, H_i),
\end{equation}
where $H_i$ denotes the hidden states of the corresponding DiT Block at layer $i$, 
serving as keys and values in the cross-attention. The output of each Kinematic 
Block is additively injected into the corresponding frozen DiT Block via a learned 
gating projection $W_g$:
\begin{equation}
    H_{i+1}^{\mathrm{main}} = H_{i+1}^{\mathrm{main}} + W_g \cdot \mathcal{K}_{i+1},
\end{equation}
where $W_g$ is initialized to zero so that the kinematic injection has no effect 
at the start of training, gradually learning to incorporate geometric guidance 
while preserving the pretrained generative capacity of the frozen DiT backbone. 
Only the Kinematic Fusion Embedder, Alignment Embedder, and Kinematic Blocks 
are trainable. All DiT parameters remain frozen throughout.

\subsection{Kinematics-Guided Video Diffusion}

IOI builds upon a pretrained latent-space video diffusion model based on flow matching as the generative backbone, into which kinematic priors are injected via the MKAI module. Temporal observation history and future target frames are first mapped into latent representations using a pretrained video Variational Autoencoder:
\begin{equation}
    z^{h} = \mathrm{Enc}(V_{t-h:t}), \qquad 
    z^{f} = \mathrm{Enc}(\hat{V}_{t+1:t+T}),
\end{equation}
where $V_{t-h:t}$ denotes the observed historical frames serving as scene context, and $\hat{V}_{t+1:t+T}$ denotes the future frames to be generated, which are available as ground truth during training and recovered via denoising during inference.

Following standard practice in interactive world model, we only apply the flow matching objective to the future latents $z^f$, while the historical observation latents $z^h$ are kept fixed as conditioning context throughout the denoising process. Specifically, we define a linear interpolation between a Gaussian noise prior $\epsilon \sim \mathcal{N}(0, I)$ and the target future latent $z^f$. For a flow time $\tau \in [0, 1]$, the noised latent $z_\tau$ is defined as:
\begin{equation}
    z_\tau = (1 - \tau) \, \epsilon + \tau \, z^{f}.
\end{equation}

The corresponding target velocity field, which transports the sample from the prior to the data distribution, is analytically given by $u_\tau = z^{f} - \epsilon$. The neural network approximates this conditional vector field, conditioned on the historical latents and the aggregated kinematic tokens:
\begin{equation}
    v_{\theta} = v_{\theta}\bigl(z_\tau, \tau;\, z^{h}, C^{\mathrm{agg}}\bigr).
\end{equation}

The model is trained by minimizing the flow matching loss:
\begin{equation}
    \mathcal{L}_{\mathrm{FM}} 
    = \mathbb{E}_{z^{f}, \epsilon, \tau}
    \left[
        \bigl\| (z^{f} - \epsilon) - v_{\theta}(z_\tau, \tau;\, z^{h}, C^{\mathrm{agg}}) \bigr\|_2^2
    \right].
\end{equation}

This objective guides the network to reconstruct future video trajectories while remaining explicitly conditioned on the robot's deterministic kinematic prior. Since embodiment motion is fully absorbed by the MKAI-injected geometric guidance, the generative model can dedicate its representational capacity to modeling the stochastic aspects of scene dynamics, such as contact-induced object interactions and non-rigid environmental responses.

\section{Experiments}
\label{sec:experiments}

\begin{table*}[t]
  \centering
  \caption{Quantitative comparison of kinematic fidelity and physical compliance on RoboTwin 2.0 under expert success trajectories. Best results are highlighted in bold, and second-best are \underline{underlined}. $\uparrow$ ($\downarrow$) indicates that higher (lower) values are preferred. IOI achieves state-of-the-art performance across the vast majority of manipulation tasks and evaluation metrics.}
  \label{tab:level1_kinematic_simple}
  \resizebox{\textwidth}{!}{ 
  \begin{tabular}{l cccc cccc cccc}
    \toprule
    \multirow{2}{*}{Task} & \multicolumn{4}{c}{IRASim~\cite{zhu2025irasim}} & \multicolumn{4}{c}{Ctrl-World~\cite{guo2026ctrlworld}} & \multicolumn{4}{c}{\textbf{IOI (Ours)}} \\
    \cmidrule(lr){2-5} \cmidrule(lr){6-9} \cmidrule(lr){10-13}
    & PSNR $\uparrow$ & SSIM $\uparrow$ & LPIPS $\downarrow$ & FVD $\downarrow$ 
    & PSNR $\uparrow$ & SSIM $\uparrow$ & LPIPS $\downarrow$ & FVD $\downarrow$ 
    & PSNR $\uparrow$ & SSIM $\uparrow$ & LPIPS $\downarrow$ & FVD $\downarrow$ \\
    \midrule
    
    Adjust Bottle 
    & \textbf{23.77} & 0.7610 & 0.1223 & 400.39 
    & 22.67 & 0.7654 & 0.1163 & \textbf{284.53}
    & 23.00 & \textbf{0.8015} & \textbf{0.1004} & 344.83 \\


    \multicolumn{1}{c}{\raisebox{0.5ex}{\textbf{\dots}}} & & & & & & & & & & & & \\
    Click Alarmclock
    & 28.87 & 0.8589 & 0.0589 & 348.47
    & 28.36 & 0.8665 & 0.0619 & 244.43 
    & \textbf{29.00} & \textbf{0.9122} & \textbf{0.0435} & \textbf{173.30} \\

    Move Can Pot
    & 29.66 & 0.8678 & 0.0496 & 236.33 
    & 28.19 & 0.8637 & 0.0606 & 214.81 
    & \textbf{30.41} & \textbf{0.9288} & \textbf{0.0310} & \textbf{89.43} \\
    
    Place A2B Left
    & 24.14 & 0.7718 & 0.1087 & 445.12 
    & 22.72 & 0.7808 & 0.1053 & 294.24
    & \textbf{24.29} & \textbf{0.8195} & \textbf{0.0886} & \textbf{193.43} \\

    Place Empty Cup
    & 28.37 & 0.8287 & 0.0631 & 464.15 
    & 28.06 & 0.8337 & 0.0624 & 327.57
    & \textbf{29.07} & \textbf{0.9038} & \textbf{0.0385} & \textbf{175.29} \\

    Place Mouse Cap
    & 26.63 & 0.8153 & 0.0781 & 419.34 
    & 25.64 & 0.8184 & 0.0760 & 325.42
    & \textbf{26.83} & \textbf{0.8672} & \textbf{0.0571} & \textbf{207.33} \\

    Place Object Stand
    & 28.06 & 0.8581 & 0.0635 & 409.58 
    & 26.97 & 0.8527 & 0.0655 & 271.76
    & \textbf{28.19} & \textbf{0.9033} & \textbf{0.0481} & \textbf{230.57} \\

    Stack Bowls Three
    & 29.00 & 0.8478 & 0.0705 & 324.85 
    & 29.05 & 0.8606 & 0.0670 & 217.87
    & \textbf{29.59} & \textbf{0.8972} & \textbf{0.0521} & \textbf{178.08} \\

    Stamp Seal
    & 26.60 & 0.7987 & 0.0703 & 373.60 
    & 24.34 & 0.8012 & 0.0847 & 319.61
    & \textbf{26.64} & \textbf{0.8829} & \textbf{0.0467} & \textbf{173.35} \\
   
    
    Turn Switch 
    & \textbf{28.79} & 0.8442 & \textbf{0.0670} & 313.95 
    & 26.57 & 0.8261 & 0.1088 & 233.87 
    & 27.79 & \textbf{0.8914} & 0.0674 & \textbf{168.36} \\
    
    \midrule 
    
    \textbf{Overall} 
    & \textbf{26.81} & \underline{0.8198} & \underline{0.0803} & 126.20 
    & 25.50 & 0.8192 & 0.0867 & \underline{64.90} 
    & \underline{25.73} & \textbf{0.8637} & \textbf{0.0695} & \textbf{41.23} \\
    \bottomrule
  \end{tabular}
  }
\end{table*}

Quantitative experiments include three simulation settings in Sec.~\ref{sec:sim}, two real-world scenarios in Sec.~\ref{sec:real}, and two ablation studies in Sec.~\ref{sec:ablation}. Together, these results validate IOI as a reliable and interactive space for physical imagination, where the model accurately anticipates and visualizes the complex consequences of robotic interventions within a consistent world representation.

\subsection{Experimental Setup}
\paragraph{\textbf{Implementation Details}} Our framework is built upon the WAN 2.1 base model~\cite{wan2025wan}. The model is pre-trained on general video clip datasets, followed by domain-specific fine-tuning on either the RoboTwin 2.0 dataset or a real-world dataset. We allocate 100 trials per task for training and withhold 10 unseen trials for evaluation. The training process consists of $100,000$ steps with a global batch size of $32$, utilizing the AdamW optimizer. The learning rate features a $2,000$-step linear warm-up to a peak of $5 \times 10^{-5}$, followed by a cosine annealing decay schedule. To enhance robustness against environmental perturbations, spatial cropping and photometric jitter are applied during the fine-tuning stage.

\paragraph{\textbf{Baselines}} We compare our approach with two state-of-the-art interactive world model methods, IRASim~\cite{zhu2025irasim} and Ctrl-World~\cite{guo2026ctrlworld}. We do not include BridgeV2W~\cite{bridgev2w2025} and Kinema4D~\cite{xu2026kinema4d} as baselines, as their implementations are not publicly available at the time of writing, precluding fair and reproducible comparison.

\paragraph{\textbf{Metrics}} We evaluate synthesis quality using four standard metrics, where PSNR and SSIM quantify pixel-level accuracy and structural preservation, LPIPS assesses perceptual realism, and FVD measures spatiotemporal coherence and kinematic drift. For policy evaluation, we report Success Rate (SR) as the primary metric.

\begin{figure*}
    \centering
    \includegraphics[width=0.98\linewidth]{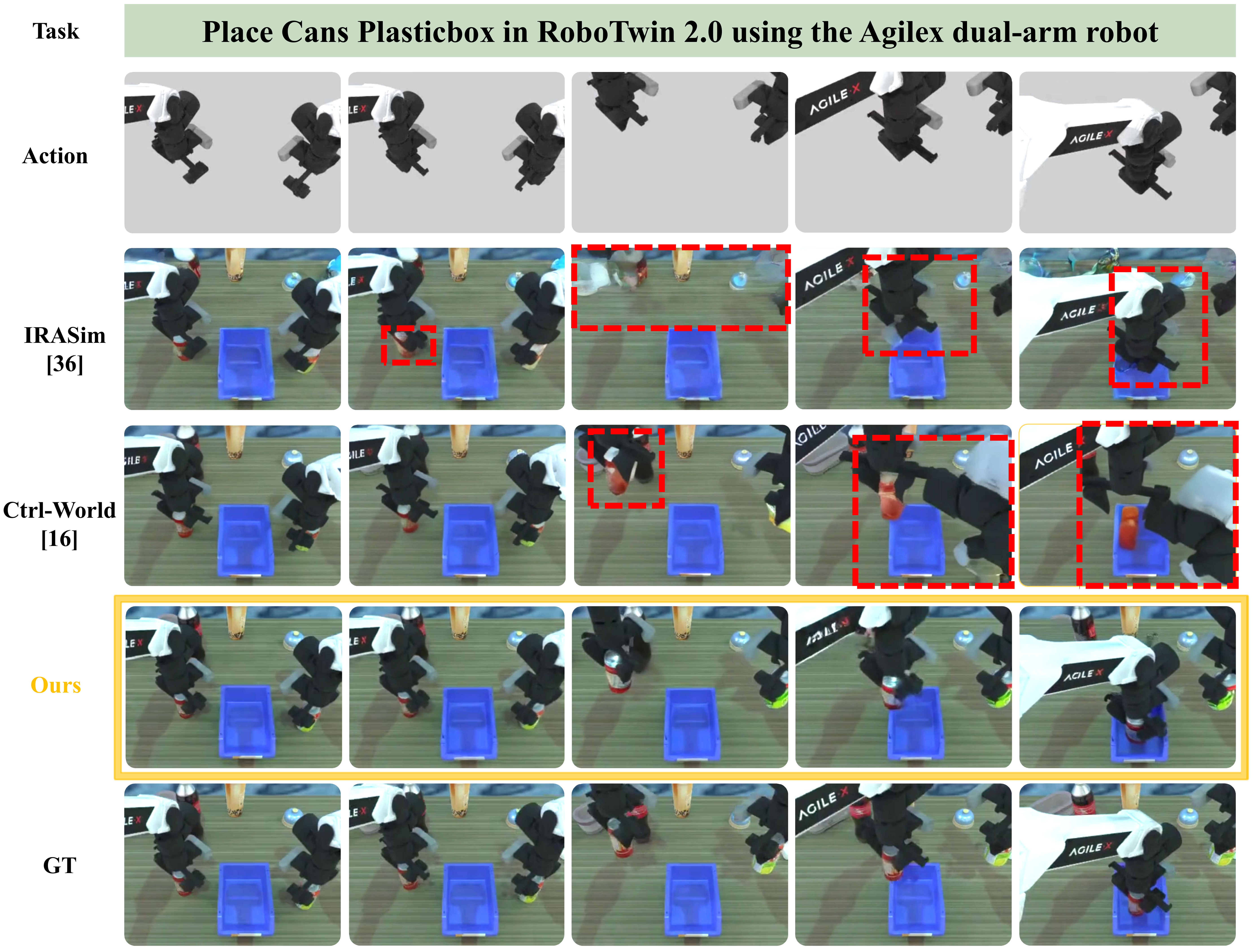}
    \caption{Visual comparison on Experiment 1. Red boxes indicate regions with significant morphological distortion or physical inconsistency in baselines, while the yellow box highlights the superior stability and realism.}
    \label{fig:visual_kinematic}
\end{figure*}

\subsection{Simulation Experiment Setup}

To evaluate the effectiveness of IOI, we design a comprehensive benchmarking suite on RoboTwin 2.0 comprising three targeted experiments.
\textbf{Experiment 1} assesses kinematic control precision and the physical compliance of robot-object interactions.
\textbf{Experiment 2} investigates the generalization capabilities of the model when confronted with out-of-distribution (OOD) tasks.
\textbf{Experiment 3} validates the utility of the model as a policy evaluator by examining its discriminative capacity between successful and failed trajectory executions.
Collectively, these three perspectives establish a holistic evaluation paradigm. They systematically cover the fundamental requirements of an embodied world model by verifying its low-level simulation fidelity, mid-level robustness to novel tasks, and high-level practical applicability for downstream policy learning.

\textbf{Experiment 1} assesses kinematic fidelity and physical compliance during dynamic robot-object interactions. We systematically benchmark the model across the entire suite of 50 manipulation tasks within RoboTwin 2.0. For each task, the model is tasked with synthesizing video sequences conditioned on initial states and unseen action trajectories. By leveraging diverse scene configurations and a wide array of interacting objects, this evaluation rigorously validates the model's capacity to maintain control precision and structural realism under complex contact dynamics.

\textbf{Experiment 2} evaluates zero-shot compositional generalization to determine the model's capacity to adapt learned physical priors to out-of-distribution (OOD) manipulation skills. We establish a strict cross-task evaluation protocol by training the model on 40 distinct tasks and reserving the remaining 10 exclusively for inference. This configuration requires the dynamic recombination of learned manipulation primitives within unfamiliar contexts, rigorously assessing whether the framework successfully decouples fundamental interaction logic from specific task semantics. Detailed lists of the specific training and evaluation tasks are provided in the Appendix.

\textbf{Experiment 3} validates the utility of the model as a reliable policy evaluator by examining its discriminative capacity between successful and failed executions. Leveraging a balanced training distribution of success and failure rollouts, the framework autoregressively simulates the $\pi_0$ vision-language-action (VLA) policy~\cite{black2024pi0} across five distinct manipulation tasks (\textit{press stapler}, \textit{block ranking rgb}, \textit{click alarmclock}, \textit{grab roller}, and \textit{adjust bottle}), deliberately selected to represent varying levels of execution success rates. We execute 50 generated action sequences per task within both our world model and the ground-truth reference simulator. Crucially, the success rates for the synthesized rollouts are determined through a rigorous double-blind cross-evaluation by three independent evaluators, strictly adhering to the identical task completion criteria employed by the simulator. By comparing these performance distributions, this experiment rigorously determines whether the world model faithfully replicates true policy outcomes and accurately captures fine-grained contact dynamics~\cite{gu2025worldgym, tseng2025scalable}.

\begin{table*}[t]
\centering
\caption{\textbf{Appendix: Detailed Breakdown of Compositional Generalization Tasks.} The 10 test tasks are constructed by recombining sub-skills covered by the 40 training tasks.}
\label{tab:appendix_compositional_tasks}
\resizebox{\textwidth}{!}{%
\begin{tabular}{l l l l}
\toprule
\textbf{Test Task (OOD)} & \textbf{Required Sub-skills} & \textbf{Training Coverage (Evidence)} & \textbf{Generalization Type} \\
\midrule
\texttt{place\_burger\_fries} & Food grasping, Dual-object placement, Target alignment & \texttt{place\_bread\_basket}, \texttt{place\_object\_stand} & Compositional Generalization \\
\texttt{place\_dual\_shoes} & Shoe grasping, Repetitive placement (2$\times$) & \texttt{place\_shoe}, \texttt{place\_object\_basket} & Instance Scaling (Single$\to$Dual) \\
\texttt{place\_cans\_plasticbox} & Can grasping, Multi-object container insertion & \texttt{place\_can\_basket}, \texttt{pick\_dual\_bottles} & Container + Multi-instance \\
\texttt{pick\_diverse\_bottles} & Bottle ID \& Grasp, Cross-appearance transfer & \texttt{pick\_dual\_bottles}, \texttt{adjust\_bottle} & Appearance Transfer \\
\texttt{dump\_bin\_bigbin} & Container grasp, Dumping motion, Large bin alignment & \texttt{put\_bottles\_dustbin}, \texttt{move\_*} & Action Chain Extension \\
\texttt{move\_can\_pot} & Can grasping, Target transfer, Pot-object relation & \texttt{move\_pillbottle\_pad}, \texttt{lift\_pot} & Object Composition \\
\texttt{handover\_mic} & Handover primitive, End-effector docking & \texttt{handover\_block} & Object Substitution \\
\texttt{rotate\_qrcode} & Planar pose adjustment, Target orientation rotation & \texttt{adjust\_bottle}, \texttt{turn\_switch} & Rotation Precision \\
\texttt{open\_microwave} & Opening (pull/lift), Handle interaction, Joint constraints & \texttt{open\_laptop} & Mechanism Migration \\
\texttt{shake\_bottle\_horizontally} & Shaking primitive, Direction constraint (Horizontal) & \texttt{shake\_bottle} & Directional Conditioning \\
\bottomrule
\end{tabular}%
}
\end{table*}

\subsection{Simulation Experiment Results}
\label{sec:sim}

Table~\ref{tab:level1_kinematic_simple} presents the quantitative comparison of \textbf{Experiment 1}. More Details are listed in Table~\ref{tab:level1_kinematic_simple}. IOI achieves state-of-the-art performance in temporal coherence, structural preservation, and perceptual realism, yielding the best FVD (41.23), SSIM (0.8637), and LPIPS (0.0695). Notably, IOI reduces FVD by 36.47\% and 67.30\% compared to Ctrl-World (64.90) and IRASim (126.20), respectively. This substantial improvement demonstrates that our explicit geometric scaffolding effectively mitigates the temporal control drift prevalent in purely generative models. While IRASim obtains a marginally higher single-frame PSNR score (26.81), it suffers catastrophic degradation in temporal coherence, indicating severe overfitting to static pixel-level synthesis at the expense of dynamic consistency. In contrast, IOI successfully reconciles high perceptual quality with kinematic precision, ensuring physically plausible long-horizon generation.

Qualitative comparisons in Figure~\ref{fig:visual_kinematic} visually corroborate these numerical advantages through a complex dual-arm manipulation sequence featuring an AgileX robot placing cans into a plastic box. IRASim exhibits catastrophic structural collapse during the rollout. The robot arms vanish entirely in intermediate frames and suffer from extreme morphological distortion subsequently, confirming its vulnerability to temporal drift. Furthermore, Ctrl-World struggles with fine-grained physical interactions, demonstrating notable object hallucination and incorrect spatial boundaries between the end-effector and the target can during the grasping phase. Conversely, IOI consistently maintains rigid robot morphology and precise hand-object interaction dynamics throughout the entire long-horizon execution. By effectively leveraging explicit action conditioning, our method successfully avoids unphysical visual artifacts and closely mirrors the ground-truth simulation trajectory.

Table~\ref{tab:appendix_compositional_tasks} details the 10 OOD test cases. For each task, we specify: (i) the decomposed sub-skills required for successful execution; (ii) the exact training tasks that provide empirical coverage for each sub-skill; and (iii) the targeted generalization category. For instance, \texttt{place\_burger\_fries} is OOD because it requires simultaneously applying a deformable-object grasp policy (learned from \texttt{place\_bread\_basket}) with a dual-target placement and alignment constraint (learned from \texttt{place\_object\_stand}), a combination never encountered during training. The test suite systematically spans seven compositional dimensions: instance scaling, appearance/cross-modal transfer, container-multi-object coupling, action chain extension, object substitution, rotational precision, and mechanism migration.

\begin{table}[h]
  \centering
  \caption{Quantitative evaluation of zero-shot OOD task generalization. Superior performance on unseen tasks demonstrates the universal applicability and generalization of IOI.}
  \label{tab:level2_ood}
  \begin{tabular}{lcccc}
    \toprule
    Method & PSNR $\uparrow$ & SSIM $\uparrow$ & LPIPS $\downarrow$ & FVD $\downarrow$ \\
    \midrule
    IRASim~\cite{zhu2025irasim} & 24.02 & 0.7740 & 0.1521 & 166.14 \\
    Ctrl-World~\cite{guo2026ctrlworld}        & 21.64 & 0.7672 & 0.1771 & 78.79 \\
    \textbf{IOI (Ours)}         & \textbf{24.08} & \textbf{0.8127} & \textbf{0.1309} & \textbf{69.16} \\
    \bottomrule
  \end{tabular}
\end{table}

Quantitative results for \textbf{Experiment 2} in Table~\ref{tab:level2_ood} demonstrate that IOI maintains strong generation quality on unseen OOD tasks, achieving the best performance across all four metrics with FVD (69.16), PSNR (24.08), SSIM (0.8127), and LPIPS (0.1309). Notably, IOI outperforms Ctrl-World by 12.23\% in FVD and IRASim by 58.39\%, demonstrating that the explicit kinematic prior generalizes effectively to novel manipulation tasks without task-specific fine-tuning. Unlike IRASim, which degrades significantly in temporal coherence and perceptual quality despite comparable pixel-level reconstruction, IOI maintains physically consistent dynamics even in unseen task configurations. Similarly, while Ctrl-World achieves competitive PSNR on in-distribution tasks, its performance degrades notably under OOD conditions, suggesting that purely data-driven methods struggles to generalize beyond the training task distribution.

\begin{table}[h]
\centering
\caption{Comparison of $\pi_0$ policy success rates evaluated in the ground truth (GT) simulator and our IOI world model across diverse manipulation tasks.}
\label{tab:policy_eval}
\renewcommand{\arraystretch}{1.1}
\small
\begin{tabular}{lccc}
\toprule
\textbf{Task} & \textbf{GT Sim. (\%)} & \textbf{IOI (Ours) (\%)} & \textbf{$\Delta$ Diff.} \\
\midrule
Adjust Bottle      & 96.0 & 96.0 & 0.0  \\
Blocks Ranking RGB & 12.0 & 14.0 & +2.0 \\
Click Alarmclock   & 24.0 & 24.0 & 0.0  \\
Grab Roller        & 82.0 & 86.0 & +4.0 \\
Press Stapler      & 6.0  & 8.0  & +2.0 \\
\midrule
\textbf{Average}   & \textbf{44.0} & \textbf{45.6} & \textbf{+1.6} \\
\bottomrule
\end{tabular}
\end{table}
Quantitative results for \textbf{Experiment 3} in Table~\ref{tab:policy_eval} show that success rates evaluated by IOI highly align with the ground-truth simulator across all task complexities. IOI correctly identifies difficult tasks with similarly low success rates, matches easy tasks with nearly identical high scores, and maintains an extremely small average difference of only +1.6 percentage points. Without hallucination-based inflation, IOI accurately models physical constraints and fine-grained contact dynamics. 
It achieves a Pearson correlation coefficient of $0.9992$ with ground truth simulator evaluations, demonstrating strong alignment between IOI-based and physical evaluation outcomes.

\subsection{Real-World Experiment Setup and Results}
\label{sec:real}

\begin{figure*}[t]
  \centering
  \includegraphics[width=\textwidth]{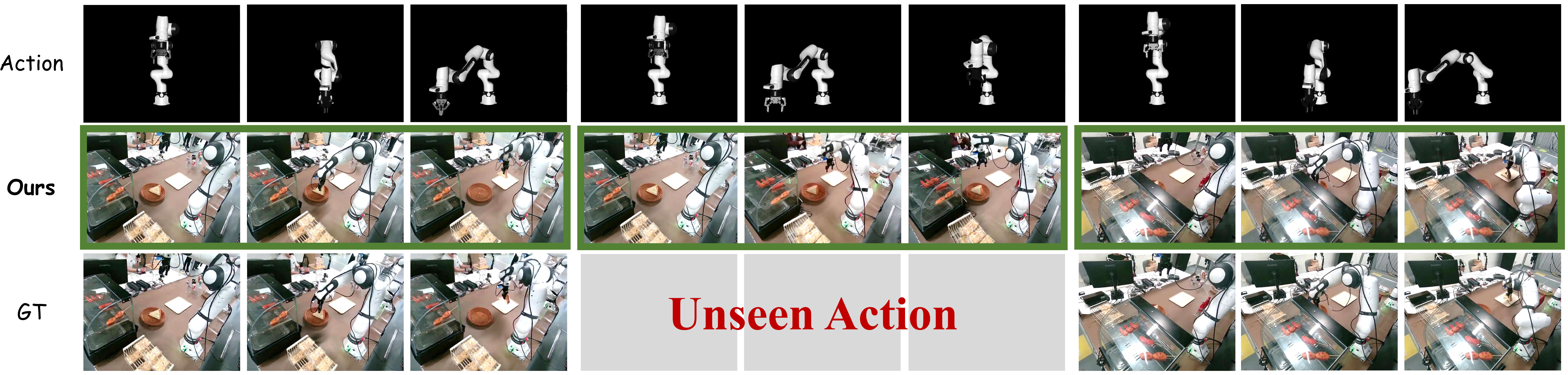}
  \caption{Qualitative results on real-world platforms demonstrate the effectiveness of IOI across diverse scenarios. Samples 1 and 3 demonstrate accurate action following under the same task with different initial states, while Sample 2 shows successful generalization to novel, unseen robotic manipulation actions, all with high visual fidelity and temporal consistency.}
  \label{fig:real}
\end{figure*}

To validate the practical applicability of IOI beyond simulation, we conduct experiments on a physical Franka Emika manipulator. The evaluation task involves 
\textit{picking a sandwich from a plate and placing it into a basket}, representing a contact-rich manipulation scenario with realistic object interactions. For this task, we collect a set of 
successful teleoperation demonstrations as training data, which serve as the visual context and action sequences for IOI.

Figure~\ref{fig:real} presents qualitative visualizations of real-world video generation 
on the sandwich pick-and-place task. IOI successfully maintains the rigid structural 
integrity of the robotic arm and synthesizes precise hand-object contact dynamics across 
diverse action sequences. Even under unstructured physical lighting conditions and 
varying object textures, the generated trajectories exhibit no severe visual distortions 
or temporal drift. Notably, although trained only on successful demonstrations of this 
specific task, IOI exhibits emergent generalization to complex and unseen action 
trajectories, suggesting that the explicit kinematic prior enables the model to capture 
underlying physical interaction patterns rather than merely memorizing demonstrated 
motions. Due to space constraints, comprehensive quantitative metrics and further 
analyses are provided in the Appendix.

\subsection{Ablation Study}
\label{sec:ablation}

\noindent\textbf{Analysis of Action Conditioning Mechanisms.} Following the evaluation protocol of Simulation \textbf{Experiment 1}, we conduct an ablation study using simulator-rendered sequences as ground-truth references to compare our explicit geometric formulation, URDF combined with MKAI, against four mainstream implicit action injection baselines. Specifically, the \textit{Add} strategy~\cite{ho2020ddpm} performs an element-wise addition of action embeddings into visual features. The \textit{Concat} method~\cite{zhu2025irasim} applies a channel-wise concatenation of action signals along the feature dimension. The \textit{Cross-attention} mechanism~\cite{rombach2022ldm} injects action representations directly through attention blocks. Finally, the \textit{AdaLN} approach~\cite{peebles2023dit} dynamically modulates the scale and shift parameters of normalization layers using action signals. This setup rigorously isolates the injection variable to objectively observe how different condition formatting directly affects the video generation process.

\begin{table}[htbp]
\centering
\caption{Quantitative ablation of action conditioning mechanisms under Simulation Experiment 1.}
\label{tab:action_ablation}
\renewcommand{\arraystretch}{1.1}
\begin{tabular}{lcccc}
\toprule
\textbf{Mechanism} & \textbf{PSNR $\uparrow$} & \textbf{SSIM $\uparrow$} & \textbf{LPIPS $\downarrow$} & \textbf{FVD $\downarrow$} \\
\midrule
Add            & 22.34 & 0.7845 & 0.1256 & 87.31 \\
Concat          & 23.58 & 0.8134 & 0.1021 & 67.89 \\
Cross-attention & 24.76 & 0.8412 & 0.0781 & 62.56 \\
AdaLN           & 24.89 & 0.8473 & 0.0812 & 56.47 \\
\midrule
\textbf{URDF + MKAI (Ours)} & \textbf{25.73} & \textbf{0.8637} & \textbf{0.0695} & \textbf{41.23} \\
\bottomrule
\end{tabular}%
\end{table}

As shown in Table~\ref{tab:action_ablation}, implicit action injection mechanisms exhibit progressively improved performance from additive injection to concatenation, cross-attention, and AdaLN, with FVD scores of 87.31, 67.89, 62.56, and 56.47 respectively, yet all remain substantially inferior to our approach. By leveraging explicit URDF kinematic priors, URDF + MKAI achieves the best performance across all metrics with PSNR of 25.73, SSIM of 0.8637, LPIPS of 0.0695, and FVD of 41.23, representing a 26.97\% FVD reduction over AdaLN. This demonstrates that explicit geometric conditioning effectively decouples deterministic embodiment motion from stochastic scene dynamics, relieving the model from implicitly inferring robot kinematics from low-dimensional action embeddings and allowing it to allocate representational capacity toward modeling complex physical interactions.

\noindent\textbf{Upper-Bound Analysis of MKAI module.} To validate the effectiveness of the Multi-view Kinematic Aggregation and Injection (MKAI) module, we evaluate its capacity to learn spatial mappings without explicit camera parameters. We construct an \textit{oracle baseline} (IOI-Explicit) within the RoboTwin simulator that utilizes ground-truth camera intrinsics and extrinsics to perfectly project the 3D skeleton onto the target pixel plane. In contrast, our proposed IOI-MKAI relies exclusively on view-invariant orthographic projections and implicitly learns the spatial alignment. We quantitatively benchmark both models on Simulation \textbf{Experiment 1}.

\begin{table}[htbp]
\centering
\caption{Quantitative upper-bound analysis of MKAI module under Simulation Experiment 1.}
\label{tab:ablation_alignment}
\renewcommand{\arraystretch}{1.1}
\begin{tabular}{llcccc}
\toprule
\textbf{Method} & \textbf{GT Extrinsics} & \textbf{PSNR $\uparrow$} & \textbf{SSIM $\uparrow$} & \textbf{LPIPS $\downarrow$} & \textbf{FVD $\downarrow$} \\
\midrule
IOI-Explicit (Oracle) & Required & 26.97 & 0.8769 & 0.0631 & 37.15 \\
\textbf{IOI-MKAI (Ours)} & \textbf{None} & 25.73 & 0.8637 & 0.0628 & 41.23 \\
\bottomrule
\end{tabular}
\end{table}

As shown in Table \ref{tab:ablation_alignment}, the IOI-Explicit baseline establishes a theoretical upper bound for generation quality by leveraging perfect spatial alignment. Remarkably, our proposed IOI-MKAI achieves highly competitive performance on the unseen test set without requiring any target camera extrinsics. This indicates that the MKAI module successfully learns the complex implicit spatial mapping from orthographic view conditions, incurring only a marginal precision trade-off. By completely eliminating the reliance on strict camera calibration, our approach significantly reduces deployment friction, enhancing the framework's robustness and generalization potential for unstructured real-world robotic applications.

\section{Conclusion}
We propose \textbf{IOI}, a hybrid interactive world model that addresses spatiotemporal drift in action-conditioned video generation by integrating analytical kinematic priors with a generative diffusion backbone. This principled decoupling of deterministic robot kinematics from stochastic scene dynamics, realized through the Multi-view Kinematic Aggregation and Injection (MKAI) module, provides view-invariant geometric guidance without requiring extrinsic camera calibration. Extensive evaluations on the RoboTwin benchmark and real-world platforms demonstrate that IOI significantly outperforms existing methods in temporal coherence, structural preservation, and perceptual realism. Its strong generalization to out-of-distribution tasks and high correlation with ground-truth policy evaluation further validate IOI as an effective and trustworthy world model for scalable embodied policy learning and evaluation. Despite these promising results, IOI currently assumes actions are specified at the joint or end-effector level, which may limit its applicability when policies operate at a higher level of abstraction. Additionally, the orthographic projection representation, while view-invariant, does not capture appearance-level details of the end-effector and grasped objects, leaving room for improvement in contact-rich interaction modeling. We hope this work encourages future exploration of physics-informed world models for embodied AI.

\clearpage

\bibliographystyle{plainnat}
\bibliography{main}




\end{document}